\documentclass{article}
 
\usepackage[acronym]{glossaries}

\usepackage{enumitem}
\usepackage{graphicx}
\graphicspath{ {./images/} }
\usepackage{subcaption}
\usepackage[dvipsnames]{xcolor}
\usepackage{lscape}
\usepackage{rotating}
\usepackage{pdflscape}
\usepackage{multirow}
\usepackage{algorithm}
\usepackage{algpseudocode}
\newtheorem{definition}{Definition}

\usepackage[final, nonatbib]{cml4impact_2022}

\usepackage[utf8]{inputenc} 
\usepackage[T1]{fontenc}    
\usepackage{hyperref}       
\usepackage{url}            
\usepackage{booktabs}       
\usepackage{amsfonts}       
\usepackage{nicefrac}       
\usepackage{microtype}      
\usepackage[dvipsnames]{xcolor}         
\usepackage{amsmath}
\usepackage{amssymb}
\usepackage{pgfplots}
\usepgfplotslibrary{groupplots,dateplot}
\usetikzlibrary{patterns,shapes.arrows}
\pgfplotsset{compat=newest}

\usepackage[
style=ieee,
sorting=none
]{biblatex}
\newcommand{\royalblue}[1]{\textcolor{RoyalBlue}{#1}}
\newcommand{\forestgreen}[1]{\textcolor{ForestGreen}{#1}}
\newcommand{\darkred}[1]{\textcolor{Red}{#1}}

\title{Causal Analysis of the \acrshort{topcat} Trial:
Spironolactone for Preserved Cardiac Function Heart
Failure}
\author{Francesca E. D. Raimondi \\
causaLens \\
\And
Tadhg O'Keeffe \\
causaLens \\
\And
Hana Chockler \\
causaLens \\
King's College London \\
\And
Andrew R. Lawrence \\
causaLens \\
\And
Tamara Stemberga \\
causaLens \\
\And
Andre Franca \\
causaLens \\
\And
Maksim Sipos \\
causaLens \\
\And
Javed Butler \\
Baylor Scott \& White Research Institute, Dallas, TX \\
University of Mississippi, Jackson, MS \\
\And
Shlomo Ben-Haim \\
Hobart Group \\
}

\makeglossaries
\newacronym{topcat}{TOPCAT}{Treatment of Preserved Cardiac Function Heart Failure with an Aldosterone Antagonist}

\newacronym{dag}{DAG}{Directed Acyclic Graph}
\newacronym{scm}{SCM}{Structural Causal Model}
\newacronym{mb}{MB}{Markov Blanket}
\newacronym{bic}{BIC}{Bayesian Information Criterion}
\newacronym{aft}{AFT}{Accelerated Failure Time}
\newacronym{smape}{SMAPE}{Symmetric Mean Absolute Percentage Error}
\newacronym{shd}{SHD}{Structural Hamming Distance}

\newacronym{rct}{RCT}{Randomized Controlled Trial}

\newacronym{cvddeath}{Outcome cvd death}{Death from cardiovascular causes - primary outcome}
\newacronym{abortedca}{Outcome abortedca}{Aborted cardiac arrest - primary outcome}
\newacronym{hfhosp}{Outcome hfhosp}{Hospitalization for hearth failure - primary outcome}

\newacronym{death}{Outcome death}{Death from any cause}
\newacronym{anyhosp}{Outcome anyhosp}{Hospitalisation for any reason}
\newacronym{mi}{Outcome mi}{Myocardial infarction}
\newacronym{stroke}{Outcome stroke}{Stroke}

\newacronym{hfpef}{HFpEF}{Heart Failure and a preserved left ventricular Ejection Fraction}

\begin{document}

\maketitle

\begin{abstract}
We describe the results of applying causal discovery methods on the data from a multi-site clinical trial, on the \acrfull{topcat}. The trial was inconclusive, with no clear benefits consistently shown for the whole cohort. However, there were questions regarding the reliability of the diagnosis and treatment protocol for a geographic subgroup of the cohort. With the inclusion of medical context in the form of domain knowledge, causal discovery is used to demonstrate regional discrepancies and to frame the regional transportability of the results. Furthermore, we show that, globally and especially for some subgroups, the treatment has significant causal effects, thus offering a more refined view of the trial results.
\end{abstract}

\section{Introduction}
Complex trials with heterogeneities in the data pose a significant challenge for traditional techniques of statistical analysis.
Suboptimal patient cohort selection and recruitment, paired with ineffective monitoring of patients, are among the main causes for high trial failure rates: only $10\%$ of the compounds entering a clinical trial reaches the market~\cite{Harrer2019:elsevier}. Phase $2$ clinical studies, albeit a poor predictor of drug success ($>30\%$ and $>58\%$ trial failure rates in phase $2$ and $3$, respectively) are critical in determining drug costs~\cite{VanNorman2019:jacc}. 
Results of multi-site clinical trials often show significant differences between the trial sites. There can be a variety of reasons for this, including different treatment benefits among different types of participants or site anomalies that could invalidate the study, such as suspected failures to assign the correct diagnosis and to adhere to treatment protocols.

The \acrshort{topcat} clinical trial is a randomized, double-blind trial: $3445$ patients with symptomatic \acrfull{hfpef} of $45\%$ or more were assigned to receive either spironolactone\footnote{Spironolactone belongs to the class of mineralocorticoid-receptor antagonists, which have been shown to improve the prognosis for patients with \acrshort{hfpef}~\cite{Pitt2014:nejm}.} ($15 mg$ to $45 mg$ daily) or placebo, with a mean follow-up of $3.3$ years. Eligible patients had a history of hospitalization within the previous $12$ months, with management of heart failure a major component of the care provided, or an elevated natriuretic peptide level within $60$ days before randomization\footnote{Brain Natriuretic Peptide (BNP) $\geq~100~pg/ml$ or N-terminal pro-BNP (NTproBNP) $\geq~360~pg/ml$.}. Trial randomization was stratified according to whether patients were enrolled on the basis of the former criterion (i.e. hospitalization stratum) or the latter criterion (i.e. the BNP stratum).
The \acrshort{topcat} clinical trial was funded by the National Heart, Lung, and Blood Institute~\cite{NHLBI2009:govct, acc:org}. Details of the trial (inclusion and exclusion criteria and covariates) and its first analysis can be found in~\cite{Pitt2014:nejm}. The goal is an evaluation of the effects of spironolactone in patients with \acrshort{hfpef}: the primary outcome is a composite of death from cardiovascular causes, aborted cardiac arrest, or hospitalization for the management of heart failure.

The trial was carried out on a geographically diverse populations with different demographic characteristics (Western vs Eastern regions)~\cite{Pfeffer2015:ahac}. Moreover, the clinical diagnostic criteria were not uniformly interpreted or applied~\cite{Aronov2014:jcec}.
While country-specific and regional heterogeneity could be viewed as expected statistical variation in a large multinational trial, the differences in patient characteristics, lower event rates, lack of certain drug side effects, and complete lack of treatment effect in Russia and Georgia compared with the Americas lead to a strong suspicion that this data is unreliable in both diagnosis and adherence to treatment~\cite{Bristow2016:jacc}. 
Transportability methods detect significant regional differences and find that measured participant characteristics do not explain the discrepancies between the sites~\cite{Berkowitz2019:ccqo}. These methods show promise in detecting anomalies within multi-center \acrfullpl{rct} and in assisting data monitoring to attribute important treatment-effect heterogeneities to participant differences or to site performance differences~\cite{Rudolph2020:epidemiology}.

Traditional methods based on correlation and survival analysis are prone to confounding and struggle to capture complex interactions: causal discovery is a necessary step in overcoming these issues.
Causality can offer new insights into clinical trials: the causal methods described in~\cite{Wu2022:nature, Dahabreh2020:sm, Prosperi2020:nature, Piccininni2020:bmc} show promise for analyzing results from \acrshortpl{rct}, via target trials, transportability, and prediction invariance.
This paper shows that causal discovery allows us to understand the causal pathways of the main outcomes of the \acrshort{topcat} trial, by deriving a causal diagram that provides insight into the regional transportability of the trial results. We were able to demonstrate regional differences of the causal mechanisms between the two main geographical subsets of the trial, offering evidence of the controversial discrepancy of the original study (Western vs Eastern countries).
Finally we show that the identified causal models can be linked to a more traditional survival analysis of the trial data, achieving better explanatory performance. 

The rest of the paper is organized as follows: the data preparation and the causal discovery are detailed in \S~\ref{sec-causal-discovery}, and the modelling of clinical outcomes following a survival analysis with the newly found causal relationships is detailed in \S~\ref{sec-survival-analysis}. A list of acronyms can be found on page \pageref{sec-acronyms}.

\section{Causal Discovery}\label{sec-causal-discovery}

A total of $3445$ participants were recruited at $233$ sites in $6$ countries ($1151$ participants in the United States, $326$ in Canada, $167$ in Brazil, $123$ in Argentina, $1066$ in Russia, and $612$ in Georgia).
The original data has a number of time varying variables based on each patient's hospital visits.
The original trial data consist of $55$ separate datasets with a total of $759$ columns. A number of these columns are identifiers such as patient ID, date of observation, etc., that are repeated across datasets. The time and patient coverage of these datasets is varied and we excluded echo-gram data due to its low coverage. There are $7$ measured outcomes and the primary outcome is a composite of $3$ of them. The outcomes are presented in a time to event format consistent with the survival analysis approach taken in the original paper \cite{Pitt2014:nejm}.

For our analysis we converted this data to a time series format to better capture the time varying nature of the data. 
We aligned the data in time and by patient to create a time series per individual for each variable present in the data.
The time resolution was set to $6$ months based on a similar average frequency of hospital visits. In each $6$ month period starting from initial randomization to $6$ years, for each patient we recorded the latest observed data and generated binary variables indicating whether a measured outcome had occurred in the given period or previously.
Baseline or static covariates were forward filled from initial randomization.
Censored data is automatically handled by this representation, as any period of survival offers a contribution to the data and the time resolution is sufficiently granular to make intra-period dropout approximately equivalent to exact time censoring. By weighting each individual equally as opposed to each row of data we also can avoid underweighting individuals with shorter survival times.

The goal of causal discovery is to identify potential causal drivers from observational data. These methods distinguish correlated variables into causal drivers and confounded variables: the latter are correlated to the target, but only because there exists a third variable, which influences both. 
\glspl{dag} are used to map all a priori assumptions surrounding a causal question and to graphically describe the underlying data generating process. Each node represents a random variable, and directed causal paths are represented by directed edges. The causal graph structure thus provides qualitative information about the conditional independence of the variables of interest, and illustrates potential sources of confounding and selection bias.

Given the complexity of the \acrshort{topcat} dataset and the importance of a few clinical outcomes, it would be inefficient to estimate the entire underlying \acrshort{dag}. Instead, we chose a feature selection strategy aimed at deriving the smallest subset of variables, excluding those that do not provide additional information on the outcomes of interest. If the lack of additional information is assessed through conditional independence tests, the problem of feature selection is analogous to identifying the \acrfull{mb} of the outcome variable.
\begin{definition}
The \acrfull{mb} of the outcome $Y$, $MB(Y)$, is the minimal subset of $X$, conditioned on which, all other variables of $X$ not included in $MB(Y)$ are independent of $Y$:
\begin{equation}
\forall V \in X \backslash MB(Y) : \quad \mathbb{P}(Y|MB(Y), V) = \mathbb{P}(Y|MB(Y))
\end{equation}
\end{definition}
If the prediction model for $Y$ can estimate the underlying true probabilities $\mathbb{P}(Y|MB(Y))$, then the variables included in the \acrshort{mb} of the outcome $Y$ are the only variables needed for an optimal prediction in terms of calibration~\cite{Piccininni2020:bmc}. 

In order to derive the \acrshort{mb} of the primary outcomes, we used PCMCI, specifically designed for time series data, which tests the variables for conditional independence for identified conditioning sets~\cite{Runge2019:aaas}. This method can work directly on the trial data converted to the time series format.
PCMCI performs two steps, consisting of an implementation of the PC algorithm, followed by the MCI algorithm. The PC Algorithm identifies a set of potential parents for both each feature and the target. After the potential parents for each variable and the target have been identified, the MCI Algorithm calculates conditional independencies conditioned on those parents. This implementation of PCMCI uses time in order to identify the causal direction. 

Our causal discovery is enhanced by the integration of medical context as a constraint structure that our algorithms have to respect. This guarantees both a better performance and time complexity. 
We specified $8$ tiers of variables, i.e. the structure of directionality that we want to enforce in our causal discovery in the form of domain knowledge for groups of variables (cf. \S~\ref{sec-dags}).
The method used for conditional independence tests is the Spearman rank-order correlation coefficient, i.e. a nonparametric measure of the monotonicity of the relationship between two datasets. Unlike the Pearson correlation, the Spearman correlation does not assume that both datasets are normally distributed and that their relationship is linear\footnote{The Spearman rank-order correlation coefficient varies between $-1$ and $+1$ with $0$ implying no correlation. Correlations of $-1$ or $+1$ imply a perfect monotone increasing and decreasing relationship, respectively.}.
We set the threshold of significance to $0.01$ for the PC step and to $0.005$ for the MCI step of the algorithm (i.e. the threshold for accepting a variable as \textit{causal} in the resulting \acrshort{dag}). 

\subsection{Causal graphs with domain knowledge}
\begin{figure}[ht!]
\centering
\includegraphics[height=0.92\textheight]{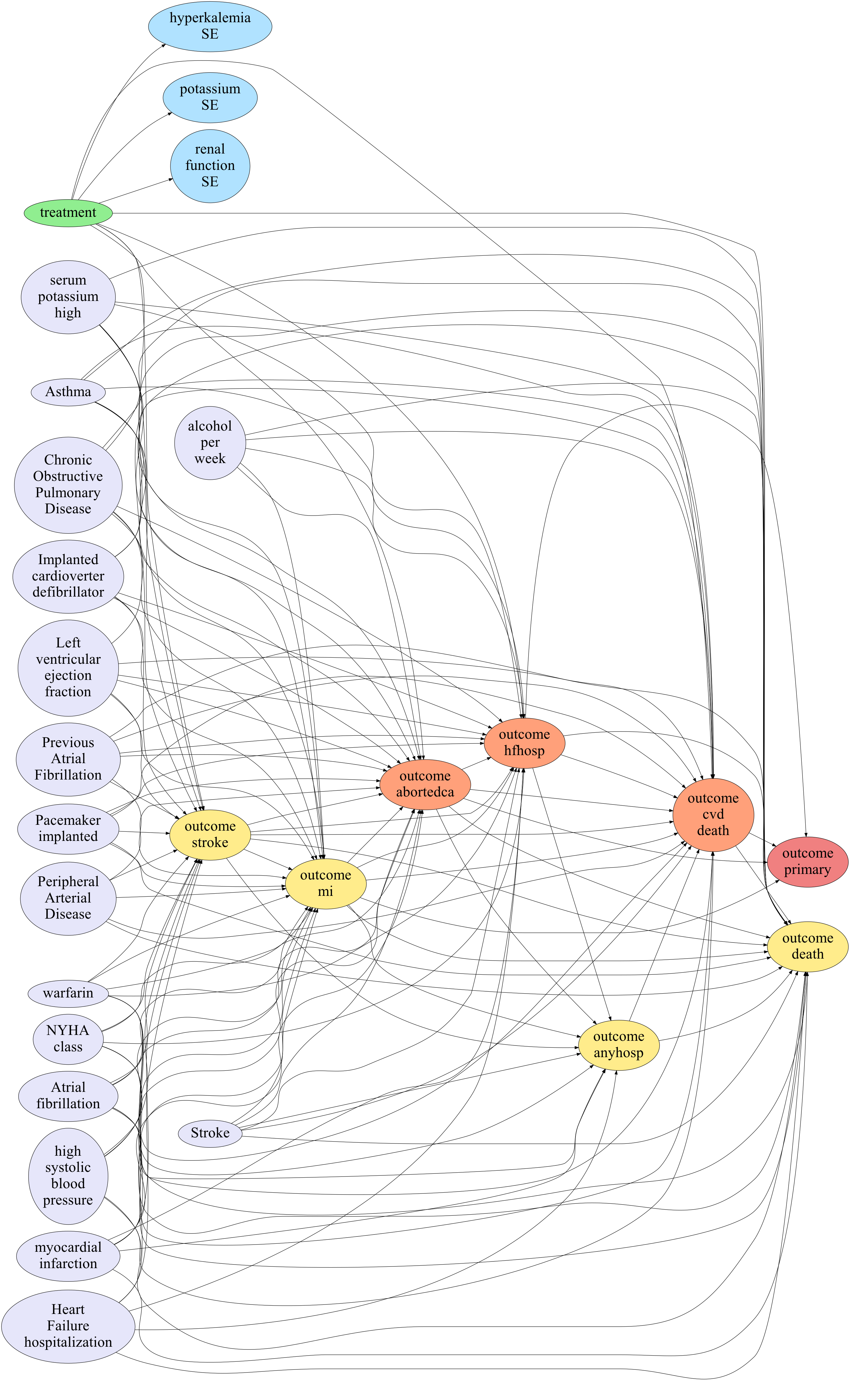}
\caption{DAG for the global dataset}
\label{fig-DAG-geo-global}
\end{figure}

\begin{figure}[ht!]
\centering
\includegraphics[width=\textwidth]{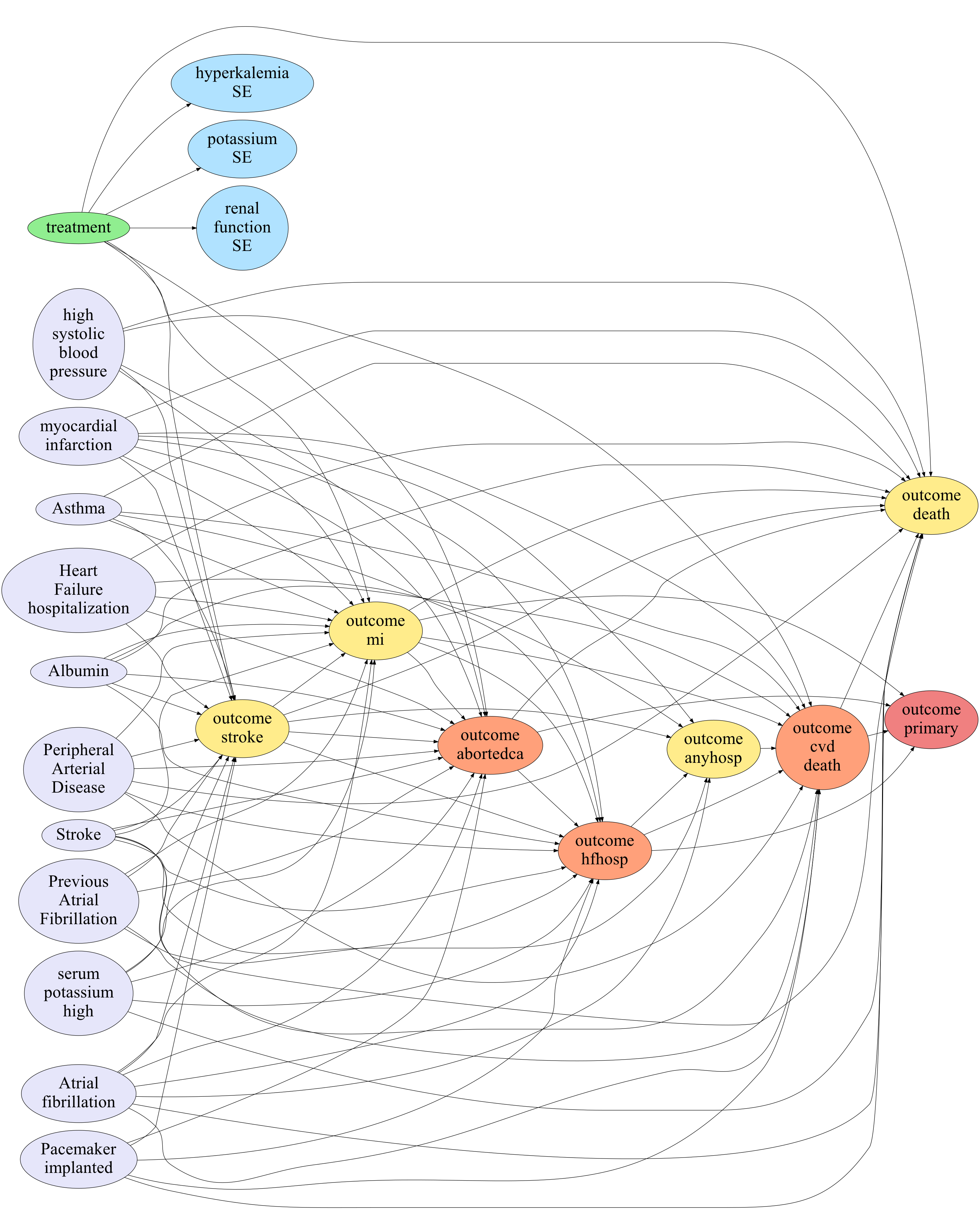}
\caption{DAG for the Western regional subset}
\label{fig-DAG-geo-regional-west}
\end{figure}

\begin{figure}[ht!]
\centering
\includegraphics[width=\textwidth]{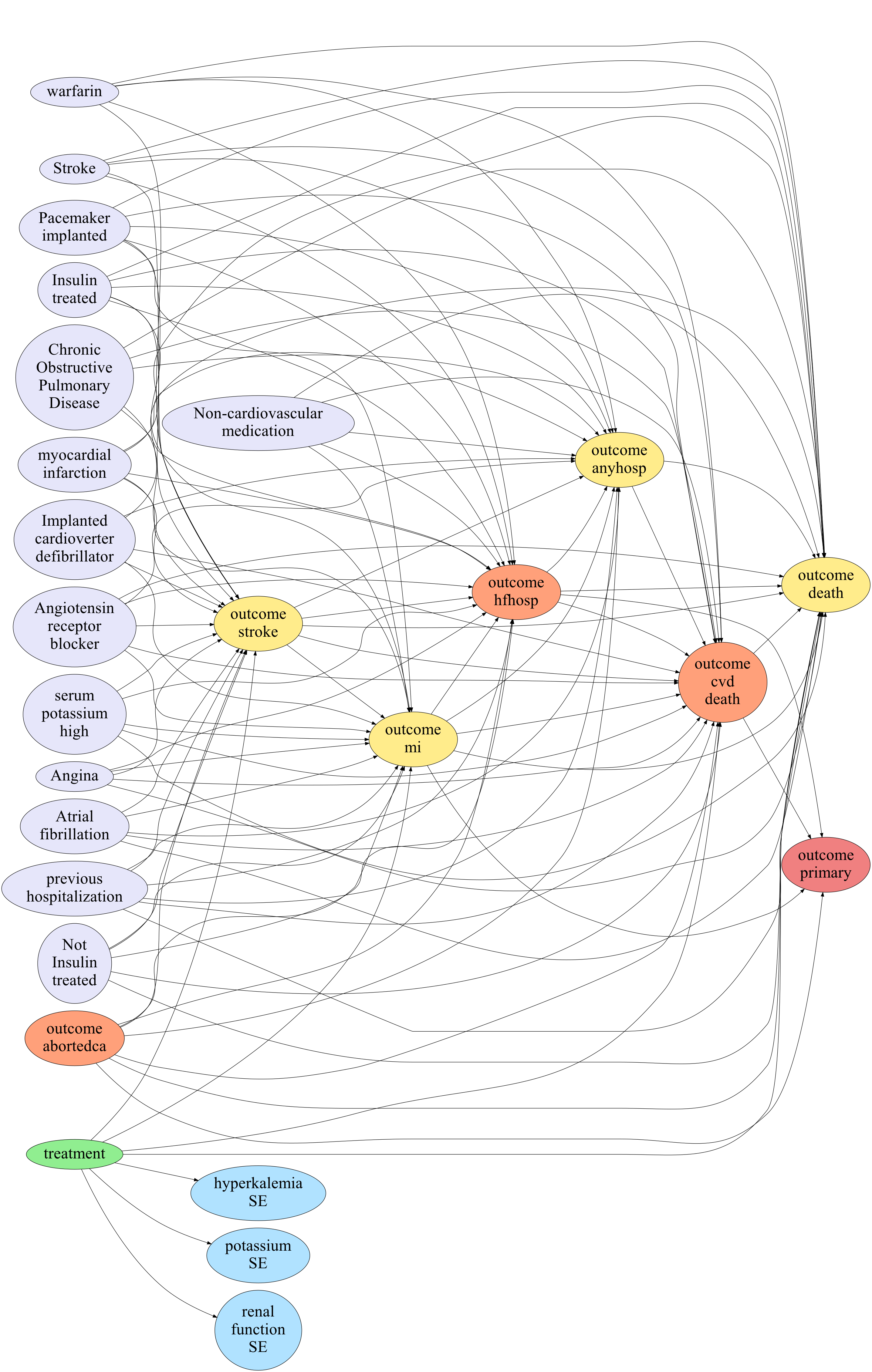}
\caption{DAG for the Eastern regional subset}
\label{fig-DAG-geo-regional-east}
\end{figure}

We applied PCMCI to achieve the causal discovery of the \acrshort{mb} of the outcomes at three scales:
\begin{enumerate}[nosep]\setlength\itemsep{0pt}
    \item Global level: the whole dataset (cf. Figures \ref{fig-DAG-geo-global} and \ref{fig-DAG-geo-global-pruned}).
    \item Regional level: the Western countries (US, Canada, Brazil, Argentina) (cf. Figures \ref{fig-DAG-geo-regional-west} and \ref{fig-DAG-geo-regional-west-pruned}).
    \item Regional level: the Eastern countries (Russia, Georgia) (cf. Figures \ref{fig-DAG-geo-regional-east} and \ref{fig-DAG-geo-regional-east-pruned}).
\end{enumerate}
We present those \acrshortpl{dag} in the supplementary material (cf. \S~\ref{sec-dags}). We also show simplified causal graphs for each scale for visualisation purposes, allowing us to focus on the identified relationships of most interest and impact by specifying a restrictive causality threshold to enforce sparsity. These diagrams visualize the causal pathways through which covariates affect outcomes through time.
We introduced additional domain knowledge of known causal pathways as edges from treatment to the known side effects and from the constituents of the primary outcome to the primary outcome event itself.

In order to explain the treatment effect, we consider the p-values of the edges that are outgoing from treatment and incoming to the outcomes (cf. Table \ref{tab-treatment-outbound-edges}).
We notice a correspondence between the global and the Western tables, both in terms of significance and of the score signs, as opposed to the Eastern table.
\footnote{The p-value and Causal Score are the result of the conditional independence test carried out by PCMCI with a significance threshold of $0.01$.}.
A negative score means that the treatment has the effect of reducing the probability of the outcomes occurring in a time period.

\begin{table}[ht!]
\small
\centering
\begin{tabular}{|c||c|c||c|c||c|c|} 
 \hline
& \multicolumn{2}{|c||}{\textbf{GLOBAL}} & \multicolumn{2}{|c||}{\textbf{WEST}} & \multicolumn{2}{|c|}{\textbf{EAST}} \\
 \hline
 \textbf{Outbound Edge} & \textbf{Score} & \textbf{P-Value} & \textbf{Score} & \textbf{P-Value} & \textbf{Score} & \textbf{P-Value} \\
\hline
\hline
 \textbf{Primary outcomes} & \multicolumn{2}{|c||}{} & \multicolumn{2}{|c||}{} & \multicolumn{2}{|c|}{} \\
  \hline
 \acrshort{cvddeath} & \forestgreen{-0.0379} & \royalblue{< 1e-04} & \forestgreen{-0.0339} & \royalblue{2.9997e-04} & \darkred{0.0432} & \royalblue{< 1e-04} \\ 
   \hline
 \acrshort{hfhosp} & \darkred{0.0353} & \royalblue{< 1e-04} & \forestgreen{-0.0250} & \royalblue{7.6480e-03} & -0.0041 & 6.3089e-01 \\
 \hline
 \acrshort{abortedca} & \darkred{0.1605} & \royalblue{< 1e-04} & \forestgreen{-0.0281} & \royalblue{2.7082e-03} & \darkred{0.0794} & \royalblue{< 1e-04}  \\
\hline
\hline
  \textbf{Secondary outcomes} & \multicolumn{2}{|c||}{} & \multicolumn{2}{|c||}{} & \multicolumn{2}{|c|}{} \\
  \hline
  \acrshort{death} & \forestgreen{-0.0827} & \royalblue{< 1e-04} & \forestgreen{-0.0502} & \royalblue{< 1e-04} & \darkred{0.0706} & \royalblue{< 1e-04} \\
 \hline
 \acrshort{anyhosp} & -0.0015 & 8.1563e-01 & -0.0007 & 9.4204e-01 & -0.0032 & 7.1122e-01 \\
 \hline
 \acrshort{mi} & \forestgreen{-0.1182} & \royalblue{< 1e-04} & \forestgreen{-0.0413} & \royalblue{< 1e-04} & \forestgreen{-0.1416} & \royalblue{< 1e-04} \\
 \hline
 \acrshort{stroke} & \forestgreen{-0.1031} & \royalblue{< 1e-04} & \forestgreen{-0.0750} & \royalblue{< 1e-04} & \forestgreen{-0.0370} & \royalblue{< 1e-04} \\
 \hline
\end{tabular}
\caption{Score and p-value of outbound edges of the node \textit{treatment}.}
\label{tab-treatment-outbound-edges}
\end{table}

Using Cox regression, the treatment was only significantly associated with \acrshort{hfhosp} in the global dataset (for results of the original study see \S~\ref{sec-original-study}). 
By modelling the causal dynamics and adjusting for the broader \acrshort{mb} set, our global analysis has mitigated the regional discrepancies to a greater degree, allowing us to identify significant causal effects of the treatment on the main outcomes as well as on additional relevant outcomes on the global data. (cf. Table \ref{tab-treatment-outbound-edges}).

In the global dataset we detected a positive causal effect of treatment on the risk of \acrshort{cvddeath}. 
We noticed a negative causal effect of treatment on the other primary outcomes (\acrshort{hfhosp} and \acrshort{abortedca}). This discrepancy with the Western subset might be a consequence of the conditioning set in the causal discovery that needs to be investigated and addressed.
We detected a positive effect of treatment on most secondary outcomes. \\
In the Western subset we detected a positive causal effect of treatment on almost all the considered primary and secondary outcomes. \\
In the Eastern subset we noticed fewer significant causal drivers in general, and more discrepancies in the signs of the interactions.

\subsection{Bootstrap analysis}
We propose a bootstrap analysis as a baseline for evaluating the regional heterogeneities, by comparison with random partitions of the data.
When examining the observed differences in causal scores and graph structures across regions, it is important to contextualize these numbers. With random variation in the data some level of difference would be expected and we can quantify this expected variation using bootstrap sampling. By drawing random samples of data partitions that are uncorrelated to Region or other covariates, we can observe the expected distribution of the differences due to randomness and compare them to the observed regional behavior. We drew $100$ of these bootstrap samples, randomly partitioning the data into two parts. After running causal discovery on all the sets of subsections, we calculated a number of metrics to capture the variation in the identified causal structures in Table \ref{tab-bootstrap-regional-differences}. 

\begin{table}[ht!]
\small
\centering
\begin{tabular}{|c|c|c|c|c|} 
 \hline
 \textbf{Measure} & \textbf{SMAPE Treatment} & \textbf{SMAPE All} & \textbf{SHD} & \textbf{Contradictions} \\
 \hline
 Observed Regional Measure & 1.500 & 1.560 & 169 & 26 \\
 \hline
 Observed quantile & $86\%$ & $100\%$ & $97\%$ & $100\%$ \\
 \hline
 \hline
 Bootstrap $25\%$ quantile &  1.044 & 1.532 & 112 & 7 \\
 \hline
 Bootstrap $50\%$ quantile &  1.223 & 1.536 & 129.5 & 9.5 \\
 \hline
 Bootstrap $75\%$ quantile & 1.401 & 1.541 & 144 & 12 \\
 \hline
\end{tabular}
\caption{Comparison of Observed Regional Score with Bootstrap Distribution.}
\acrshort{smape}\protect\footnotemark 
Treatment: \acrshort{smape} between causal scores of the treatment on each outcome. \\
\acrshort{smape} All: \acrshort{smape} between the causal scores of every covariate on each outcome. \\
\acrfull{shd}: between the graphs discovered on each subset. \\
Contradictions: the number of existing causal edges with opposite signs.
\label{tab-bootstrap-regional-differences}
\end{table}

\footnotetext{\acrfull{smape}.}

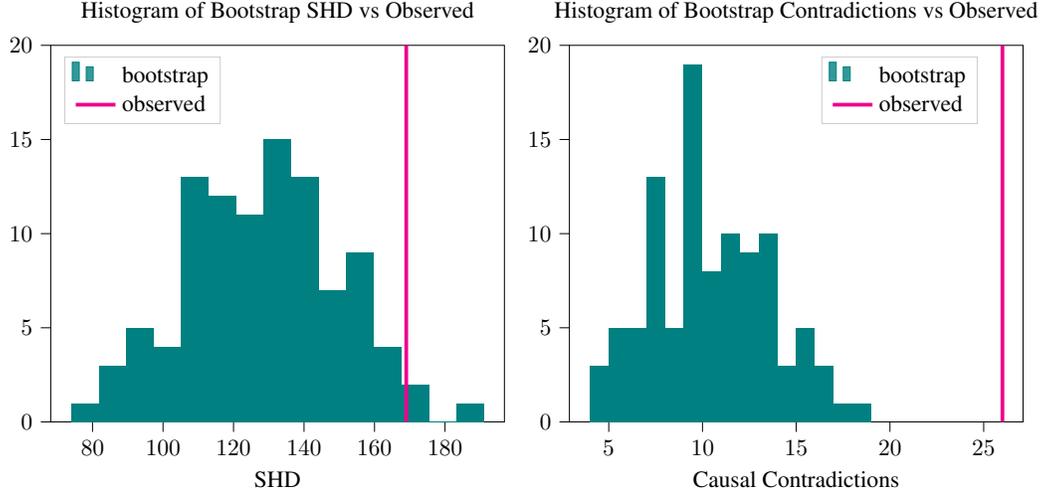
\begin{figure}[!ht]
\begin{tikzpicture}[scale=0.88]

\definecolor{darkgray176}{RGB}{176,176,176}
\definecolor{lightgray204}{RGB}{204,204,204}

\begin{axis}[
legend cell align={left},
legend style={fill opacity=0.8, draw opacity=1, text opacity=1, draw=lightgray204},
tick align=outside,
tick pos=left,
title={Histogram of Bootstrap SHD vs Observed},
x grid style={darkgray176},
xlabel={SHD},
xmin=68.15, xmax=196.85,
xtick style={color=black},
y grid style={darkgray176},
ymin=0, ymax=20,
ytick style={color=black},
legend style={
  fill opacity=0.8,
  draw opacity=1,
  text opacity=1,
  at={(0.03,0.97)},
  anchor=north west,
  draw=lightgray204
}
]

\draw[draw=none,fill=teal] (axis cs:74,0) rectangle (axis cs:81.8,1);
\draw[draw=none,fill=teal] (axis cs:81.8,0) rectangle (axis cs:89.6,3);
\draw[draw=none,fill=teal] (axis cs:89.6,0) rectangle (axis cs:97.4,5);
\draw[draw=none,fill=teal] (axis cs:97.4,0) rectangle (axis cs:105.2,4);
\draw[draw=none,fill=teal] (axis cs:105.2,0) rectangle (axis cs:113,13);
\draw[draw=none,fill=teal] (axis cs:113,0) rectangle (axis cs:120.8,12);
\draw[draw=none,fill=teal] (axis cs:120.8,0) rectangle (axis cs:128.6,11);
\draw[draw=none,fill=teal] (axis cs:128.6,0) rectangle (axis cs:136.4,15);
\draw[draw=none,fill=teal] (axis cs:136.4,0) rectangle (axis cs:144.2,13);
\draw[draw=none,fill=teal] (axis cs:144.2,0) rectangle (axis cs:152,7);
\draw[draw=none,fill=teal] (axis cs:152,0) rectangle (axis cs:159.8,9);
\draw[draw=none,fill=teal] (axis cs:159.8,0) rectangle (axis cs:167.6,4);
\draw[draw=none,fill=teal] (axis cs:167.6,0) rectangle (axis cs:175.4,2);
\draw[draw=none,fill=teal] (axis cs:175.4,0) rectangle (axis cs:183.2,0);
\draw[draw=none,fill=teal] (axis cs:183.2,0) rectangle (axis cs:191,1);

\path [draw=magenta, ultra thick]
(axis cs:169,0)
--(axis cs:169,20);

\addlegendimage{ybar,ybar legend,draw=teal,fill=teal}
\addlegendentry{bootstrap}
\addlegendimage{draw=magenta, ultra thick}
\addlegendentry{observed}

\end{axis}

\end{tikzpicture}
\hfill
\begin{tikzpicture}[scale=0.88]

\definecolor{darkgray176}{RGB}{176,176,176}
\definecolor{lightgray204}{RGB}{204,204,204}

\begin{axis}[
legend cell align={left},
legend style={fill opacity=0.8, draw opacity=1, text opacity=1, draw=lightgray204},
tick align=outside,
tick pos=left,
title={Histogram of Bootstrap Contradictions vs Observed},
x grid style={darkgray176},
xlabel={Causal Contradictions},
xmin=2.9, xmax=27.1,
xtick style={color=black},
y grid style={darkgray176},
ymin=0, ymax=20,
ytick style={color=black},
legend style={
  fill opacity=0.8,
  draw opacity=1,
  text opacity=1,
  at={(0.90,0.97)},
  anchor=north east,
  draw=lightgray204
}
]

\draw[draw=none,fill=teal] (axis cs:4,0) rectangle (axis cs:5,3);
\draw[draw=none,fill=teal] (axis cs:5,0) rectangle (axis cs:6,5);
\draw[draw=none,fill=teal] (axis cs:6,0) rectangle (axis cs:7,5);
\draw[draw=none,fill=teal] (axis cs:7,0) rectangle (axis cs:8,13);
\draw[draw=none,fill=teal] (axis cs:8,0) rectangle (axis cs:9,5);
\draw[draw=none,fill=teal] (axis cs:9,0) rectangle (axis cs:10,19);
\draw[draw=none,fill=teal] (axis cs:10,0) rectangle (axis cs:11,8);
\draw[draw=none,fill=teal] (axis cs:11,0) rectangle (axis cs:12,10);
\draw[draw=none,fill=teal] (axis cs:12,0) rectangle (axis cs:13,9);
\draw[draw=none,fill=teal] (axis cs:13,0) rectangle (axis cs:14,10);
\draw[draw=none,fill=teal] (axis cs:14,0) rectangle (axis cs:15,3);
\draw[draw=none,fill=teal] (axis cs:15,0) rectangle (axis cs:16,5);
\draw[draw=none,fill=teal] (axis cs:16,0) rectangle (axis cs:17,3);
\draw[draw=none,fill=teal] (axis cs:17,0) rectangle (axis cs:18,1);
\draw[draw=none,fill=teal] (axis cs:18,0) rectangle (axis cs:19,1);
\path [draw=magenta, ultra thick]
(axis cs:26,0)
--(axis cs:26,20);

\addlegendimage{ybar,ybar legend,draw=teal,fill=teal}
\addlegendentry{bootstrap}
\addlegendimage{draw=magenta, ultra thick}
\addlegendentry{observed}

\end{axis}

\end{tikzpicture}
\caption{Bootstrap histograms: \acrshort{shd} on the left, contradictions on the right.}
\label{fig-hist}
\end{figure}

Histograms of \acrshort{shd} and the number of contradictions can be found in Figure \ref{fig-hist}, and details of the differences among single outcomes can be found in Table \ref{tab-bootstrap-regional-differences-per-outcome}.

\begin{table}[ht!]
\small
\centering
\begin{tabular}{|c||c|c||c|c|} 
 \hline
 \textbf{Outcome} & \textbf{Treatment Difference} & \textbf{Quantile} & \textbf{Mean Difference} & \textbf{Quantile} \\
 \hline
 Outcome cvd death & 0.077 & $57\%$ & 0.086 & $99\%$ \\ 
 \hline
 \acrshort{abortedca} & 0.107 & $53\%$ & 0.077 & $96\%$ \\
 \hline
 \acrshort{hfhosp} & 0.021 & $52\%$ & 0.071 & $100\%$ \\
 \hline
 \acrshort{death} & 0.121 & $80\%$ & 0.083 & $98\%$ \\
 \hline
 \acrshort{anyhosp} & 0.002 & $11\%$ & 0.038 & $100\%$ \\
 \hline
 \acrshort{mi} & 0.100 & $79\%$ & 0.070 & $100\%$ \\
 \hline
 \acrshort{stroke} & 0.038 & $30\%$ & 0.056 & $99\%$ \\
 \hline
\end{tabular}
\caption{Comparison of Observed Regional Score with Bootstrap Distribution per Outcome.}
\label{tab-bootstrap-regional-differences-per-outcome}
\end{table}

Looking at a broader causal picture has allowed us to capture regional differences very strongly when a univariate survival analysis could not.
We can see in Tables \ref{tab-bootstrap-regional-differences} and \ref{tab-bootstrap-regional-differences-per-outcome} that stratifying the data by region led to significantly higher differences in results compared to random partitions. Differences in score (\acrshort{smape}) were significant with regard to treatment ($86\%$ quantile) and extreme when taking into account other covariates ($100\%$ quantile).
Measurements around graph structure (\acrshort{shd}) also gave a clear picture ($97\%$ quantile). The number of contradictory causal conclusions between regions ($26$) was higher than any observed bootstrap sample with the highest observed being $19$ ($100\%$ quantile). This comparison strongly supports the claim of regional treatment discrepancy in terms of the difference between the corresponding causal structures. 

\section{Survival Analysis}\label{sec-survival-analysis}
We propose the use of causal discovery as an automatic tool for feature selection of the most influential covariates for a traditional survival analysis. Generally, covariates are selected manually or using correlation methods, without taking into account the causal relationship to the outcomes of interest.
In \acrshortpl{rct} with a survival outcome, treatment effects are generally estimated with the Cox hazard ratio, often without adjusting for known prognostic variables.
When interpreting Cox regression through the lens of causality, omission of confounding terms creates problems, despite randomization of treatment at the origin of time ($t=0$). Therefore, the hazard ratio does not admit a causal interpretation in the case of unknown heterogeneity~\cite{Aalen2015:springer}, as the risk sets beyond the first event time are comprised of the subset of individuals who have not previously failed. Despite randomization at $t = 0$, the contributions to the Cox partial likelihood following the first failure will not remain randomized. 

On the other hand, the effect of treatment in terms of an \acrfull{aft} model, does yield a causal measure of treatment effects as the hazard function has an additive form in the covariates~\cite{Aalen2015:springer} (cf. \S~\ref{sec-AFT} for more details).
The coefficient of the treatment variable in the \acrshort{aft} model can be interpreted as slowing down or speeding up the survival function. A coefficient $b$ changes the expected survival time by a factor of $\exp(b)$. We can see in the tables below that the Global and Western \acrshort{aft} models identify a significant effect of the treatment on heart failure hospitalization. This coefficient of $0.28$ is interpreted as meaning that the expected survival time before heart failure hospitalization is $\exp(0.28) - 1 = 32.3\%$ longer given the treatment than without. 

We present results of goodness of fit for the \acrshort{aft} regression of the outcomes in the form of \acrfull{bic}.
For large datasets, the model with minimum \acrfull{bic} ideally corresponds to the candidate model which is a posteriori most probable~\cite{Neath2012:wiley}.~ \footnote{\acrfull{bic} is based on the empirical log-likelihood $L$ and does not require the specification of priors.
The \acrshort{bic} for candidate model $M_k$ is defined as
$
\text{BIC} = -2 \ln{L(\theta_k | y)} + k \ln{n}
$
where $k$ is the number of parameters estimated by the model and $n$ is the sample size of observations $y$.
}
Results of the \acrshort{aft} regression model for the global dataset and the Western and Eastern subsets can be found in Table \ref{tab-aft}.
The rows correspond to the particular predicted outcome, whereas the columns correspond to two feature selection strategies before \acrshort{aft} modeling: (1) Features corresponding to the \acrshort{mb} of the outcomes (cf. Section \ref{sec-causal-discovery}); (2) Features used in the primary study in~\cite{Pitt2014:nejm}.
We can see that the \acrshort{mb} yielded by causal discovery in all cases leads to a lower \acrshort{bic} and correspondingly a better model of the data. This accounts for increased complexity in the model with the introduction of the new \acrshort{mb} covariates. The absolute scale of the \acrshort{bic} for each model is a function of the size of the data so does not allow for relative comparison across regions but by looking at the baseline \textit{Original model} on each region we can see relatively how well we capture the survival process in this data. 

\begin{table}[ht!]
\small
\centering
\begin{tabular}{|c||c|c|c||c|c|c||c|} 
 \hline
 \textbf{Outcome} & \textbf{Coef \acrshort{mb}} & \textbf{p-val \acrshort{mb}} & \textbf{\acrshort{bic} \acrshort{mb}} & \textbf{Coef Orig} & \textbf{p-val Orig} & \textbf{\acrshort{bic} Orig} & \\ 
 \hline
 \hline
 hfhosp & \forestgreen{0.28} & \royalblue{1.87e-02} & 3349.36 & \forestgreen{0.29} & \royalblue{2.26e-02} & 3512.87 & \multirow{4}{*}{\textbf{GLOBAL}}\\ 
 abortedca & 0.16 & 5.16e-01 & 154.17 & 0.14 & 5.67e-01 & 163.89 & \\ 
 cvd death & 0.10 & 3.67e-01 & 2997.92 & 0.11 & 3.33e-01 & 3040.80 & \\ 
\cline{1-1} \cline{4-4} \cline{7-7}
 Total & & & 6501.45 & & & 6717.56 & \\
\hline
 \hline
 hfhosp & \forestgreen{0.28} & \royalblue{4.34e-02} & 2601.16 & \forestgreen{0.30} & \royalblue{3.86e-02} & 2673.64 & \multirow{4}{*}{\textbf{WEST}} \\ 
 abortedca & 0.24 & 4.25e-01 & 106.81 & 0.24 & 4.47e-01 & 113.83 & \\ 
 cvd death & \forestgreen{0.40} & \royalblue{5.32e-03} & 1840.00 & \forestgreen{0.39} & \royalblue{7.20e-03} & 1872.07 & \\ 
\cline{1-1} \cline{4-4} \cline{7-7}
 Total & & & 4547.97 & & & 4659.54 & \\
 \hline
 \hline
 hfhosp & 0.36 & 1.57e-01 & 535.49 & 0.26 & 3.53e-01 & 609.27 & \multirow{4}{*}{\textbf{EAST}} \\ 
 abortedca & -0.02 & 9.56e-01 & 53.80 & -0.02 & 9.62e-01 & 58.99 & \\ 
 cvd death & -0.22 & 2.30e-01 & 1104.85 & -0.22 & 2.21e-01 & 1149.90& \\
\cline{1-1} \cline{4-4} \cline{7-7}
Total & & & 1694.14 & & & 1818.16 & \\
\hline
\end{tabular}
\caption{Coefficients with corresponding p-values and \acrshortpl{bic} of AFT models.}
\label{tab-aft}
\end{table}

\section{Conclusion and Perspectives}
Complex trials with heterogeneities in the data pose a significant challenge for traditional techniques of statistical analysis. 
Suboptimal patient cohort selection and recruitment, paired with ineffective monitoring of patients, are among the main causes for high trial failure rates in phases $2$ and $3$~\cite{Harrer2019:elsevier, VanNorman2019:jacc}.  
Our ambition is to use causality in order to provide earlier determination of drug futility, inform adaptive trial design, and shorten overall drug development time. 
In this paper we show that cutting edge techniques of causal discovery augmented with human domain knowledge can offer novel insights and achieve results at greater level of significance. 

Traditional correlational methods are limited by confounding and multiple pathways and traditional survival modelling is restricted to a univariate outcome, necessitating the definition of a single primary outcome.
Causal discovery allows us to understand the interaction between outcomes and covariates leading to a diagram describing the causal process of the \acrshort{topcat} dataset. 
Our causal analysis has mitigated the regional discrepancies, allowing us to identify significant global causal effects of the treatment on
the main outcomes as well as on additional relevant outcomes.
After derivation of the regional causal diagrams, our bootstrap analysis provided evidence for a different causal effect of the treatment by region.
We also applied an \acrshort{aft} modelling to the data and were able to achieve lower \acrshort{bic} after \acrshort{mb} feature selection.

As a future extension of our work, we intend to achieve better causal effect estimation and design of clinical trials, as secondary relationships are learned from this dataset via full causal discovery. A combination of more in depth domain knowledge and data driven causal inference methods could lead to better survival models.
Additional covariates could be added thus accounting for further potential confounding.
The causal framework also allows conclusions from trials to go beyond binary outcomes and further refine our analysis (e.g. including multiple levels/doses of treatment). These conclusions could be continuously derived during a clinical trial, potentially leading to changes in its framework or to an early termination.

Transportability of causal graphs~\cite{Bareinboim2012:aaai, Bareinboim2013:aaai, Bareinboim2013:ais, Petersen2011:epidemiology, Lee2020:aaai} refers to the problem of extrapolating experimental findings across domains (i.e., settings, populations, environments) that differ both in their distributions and in their inherent causal characteristics~\cite{Bareinboim2016:nas, Lesko2017:epidemiology}. For instance, this can allow us to infer the causal effect at one population from experiments conducted in a different population after noticing that some of the covariate distributions are different.  
The subject of transportability in epidemiology is also related to data fusion, i.e. the combination of compatible graphs according to their degree of confidence, similarly to~\cite{Alrajeh2020:ai}. Since the \acrshort{topcat} study combined data from multiple geographical sources that had different underlying dynamics leading to mixed results, we could combine compatible graphs from different sub-populations according to their degree of confidence. We could similarly validate data sources and our confidence in their adherence to the study design by comparing their causal dynamics to known ground truths.

\clearpage
\printbibliography

\appendix

\bigskip
\section{Supplementary Material}\label{sec-dags}

\printglossary[type=\acronymtype]\label{sec-acronyms}

\clearpage
\subsection{Tiers for causal discovery}
{\fontfamily{cmtt}\selectfont 
[

    [
        `Americas',
        `Region interaction',
        `treatment',
        `age entry',
    ],
    
    [
        `blood potassium',
        `systolic blood pressure',
        `diastolic blood pressure',
        `Left ventricular ejection fraction',
        `Heart Failure History',
        `Previous Atrial Fibrillation',
        `Known chronic hepatic disease',
        `Heart Failure hospitalization',
        `Angina',
        `Chronic Obstructive Pulmonary Disease',
        `Asthma',
        `Peripheral Arterial Disease',
        `Dyslipidemia',
        `Implanted cardioverter defibrillator',
        `Pacemaker implanted',
        `Atrial fibrillation', 
        `Coronary artery bypass graft surgery',
        `Percutaneous Coronary Revascularization',
        `Stroke',
        `myocardial infarction',
        `smoker',
        `years smoker',
        `previous smoker',
        `cigarettes smoke a day',
        `sodium intake score',
        `alcohol per week',
        `activity level',
        `weight',
        `waist circumference',
        `heart rate',
        `Use of blood pressure lowering medication',
        `Use of Non-cardiovascular medication',
        `Use of cardiac medications',
        `aspirin use',
        `nitrate use',
        `statin use',
        `warfarin use',
        `diuretic use',
        `Use of Angiotensin receptor blocker',
        `Use of Angiotensin converting enzyme inhibitor',
        `white blood cell count',
        `hemoglobin',
        `Platelet',
        `Albumin',
        `Total Bilirubin',
        `blood glucose',
        `Alkaline Phosphatase',
        `Aspartate Aminotransferase',
        `blood sodium',
        `blood chloride',
        `creatinine result',
        `gfr',
        `Insulin treated',
        `Not Insulin treated',
        `IN HOSP',
        `left atrial volume 4',
        `left atrial volume 2',
    ],
    
    [
        `serum potassium high',
        `Systolic blood pressure (SBP) $>$ 160 mm Hg',
    ],
    
    [
        `NYHA class',
        `hyperkalemia',
        `Hypertension',
    ],
    
    [
        `creatinine doubled',
        `hyperkalemia hospitalization',
    ],
    
    [
        `Outcome anyhosp',
        `Outcome mi',
        `Outcome stroke',
        `Outcome abortedca',
        `Outcome hfhosp',
    ],
    
    [
        `Outcome death',
        `\acrshort{cvddeath}',
        `hyperkalemia side effect discontinued',
        `potassium side effect discontinued',
        `renal function side effect discontinued', 
    ],
    
    [`Outcome primary ep', `target'],
    
]
}

\clearpage
\subsection{Comparison with the original study}\label{sec-original-study}
For the sake of comparison with the primary study in~\cite{Pitt2014:nejm}, we reproduce their original table of results:

\begin{table}[ht!]
\centering
\begin{tabular}{|c|c|c|} 
 \hline
 \textbf{Outcome} & \textbf{Hazard Ratio with} & \textbf{P-Value} \\
 & \textbf{Spironolactone ($95\%$ CI)} & \\
 \hline
 Primary outcome & 0.89 (0.77–1.04) & 0.14 \\
 \hline
 \textbf{Primary outcomes} & & \\
 \hline
 Outcome \acrshort{cvddeath} & 0.90 (0.73–1.12) & 0.35 \\ 
 \hline 
 \acrshort{abortedca} & 0.60 (0.14–2.50) & 0.48 \\
 \hline
 \acrshort{hfhosp} & \forestgreen{0.83 (0.69–0.99)} & \royalblue{0.04} \\
 \hline
 
 \textbf{Secondary outcomes} & & \\
 \hline
 \acrshort{death} & 0.91 (0.77–1.08) & 0.29 \\
 \hline
 \acrshort{anyhosp} & 0.94 (0.85–1.04) & 0.25 \\
 \hline
 \acrshort{mi} & 1.00 (0.71–1.42) & 0.98 \\
 \hline
 \acrshort{stroke} & 0.94 (0.65–1.35) & 0.73 \\
 \hline

\end{tabular}
\caption{Unadjusted hazard ratios calculated via Cox proportional-hazards models.}
\label{tab-cox-primary}
\end{table}

The original study in~\cite{Pitt2014:nejm} concluded that adding spironolactone to existing therapy in patients with heart failure and a preserved ejection fraction did not significantly reduce the incidence of the primary outcome (cf. Table \ref{tab-cox-primary}). 
As for the single Primary outcomes, spironolactone did not significantly reduce the composite primary end point of death from cardiovascular causes, aborted cardiac arrest, or hospitalization for heart failure.
Also, neither the time to first hospitalization for any reason nor the time to death from any cause was significantly altered by random assignment to spironolactone. These two inclusive outcome measures are important for assessing overall risk versus benefit of the drug.

Post hoc analysis indicated marked regional differences in event rates in the Supplementary Appendix of~\cite{Pitt2014:nejm} (cf Table \ref{tab-cox-region}). In the Americas (the United States, Canada, Brazil, and Argentina), the primary outcome occurred in 242 patients in the spironolactone group ($27.3\%$) and $280$ patients in the placebo group ($31.8\%$). In Russia and Georgia, the primary outcome occurred in $78$ patients in the spironolactone group ($9.3\%$) and $71$ patients in the placebo group ($8.4\%$). However, the pre-specified test for interaction between region and study group was not significant ($p = 0.12$).

\begin{table}[ht!]
\centering
\begin{tabular}{|c|c|c|} 
 \hline
 & Western subset (USA CAN ARG BRA)
 & Eastern subset (RUS GEO) \\
 & (N=1,767) & (N=1,678) \\
 \hline
\textbf{Hazard Ratio} & \forestgreen{0.82 (0.69-0.98)} & 1.10 (0.79-1.51)	\\
 \hline
 
 \textbf{P-Value} & \royalblue{0.026} & 0.576\\
 \hline
\end{tabular}
\caption{Unadjusted hazard ratios calculated via Cox proportional-hazards models by geographic region.}
\label{tab-cox-region}
\end{table}

\clearpage
\subsection{\acrfull{aft}}\label{sec-AFT}
AFT Models assume the following form:
\begin{equation}\label{eq-aft-general}
\log(T) = -\log(\theta) +\log(T\theta) = -\log(\theta) + \epsilon
\end{equation}
where $T$ denotes the time to event, $\epsilon$ is distributed according to the underlying parametric assumption, and $\theta$ denotes the joint effect of covariates $x_i$ (typically $\theta = \boldsymbol{\beta}^T \boldsymbol{x} = \exp{(\sum_i \beta_i x_i)}$). This reduces AFT modelling to a regression where we estimate the fixed effect of our covariates with the chosen parametric distribution modelling the noise. 

We used a Weibull distribution for our analysis. We can write Equation \ref{eq-aft-general} as
\begin{equation}\label{eq-aft}
\log(T) = Y = \mu + \boldsymbol{\alpha}^T\boldsymbol{x} + \sigma W
\end{equation}
where $\boldsymbol{x}$ is the set of covariates and W has extreme value distribution.
Given transformations 
\begin{equation}
\begin{cases}
\gamma = 1 / \sigma \\
\lambda = \exp{(-\mu / \sigma)} \\
\boldsymbol{\beta} = -\boldsymbol{\alpha} / \sigma
\end{cases}
\end{equation}
we obtain a Weibull model with proportional basal hazard
\begin{equation}\label{eq-weibull-hazard}
\begin{aligned}
h(t | \boldsymbol{x}) &= (\gamma \lambda t^{\gamma -1}) \exp {(\boldsymbol{\beta}^T \boldsymbol{x})} =  \\
&= h_0(t) \exp {(\boldsymbol{\beta}^T \boldsymbol{x})}
\end{aligned}
\end{equation}
Notice that the logarithm of the hazard in Equation \ref{eq-weibull-hazard} is additive in the different covariates:
\begin{equation}
\log [h(t | \boldsymbol{x})] = \log [h_0(t)] + \boldsymbol{\beta}^T \boldsymbol{x} = \log [h_0(t)] + \sum_i{\beta_i x_i}
\end{equation}

\clearpage
\subsection{\acrfullpl{dag}}

\vfill
\vfill

\clearpage
\begin{landscape}
~
\vfill
\begin{figure}[ht!]\setlength\abovecaptionskip{1cm}
\centering
\includegraphics[width=\paperwidth]{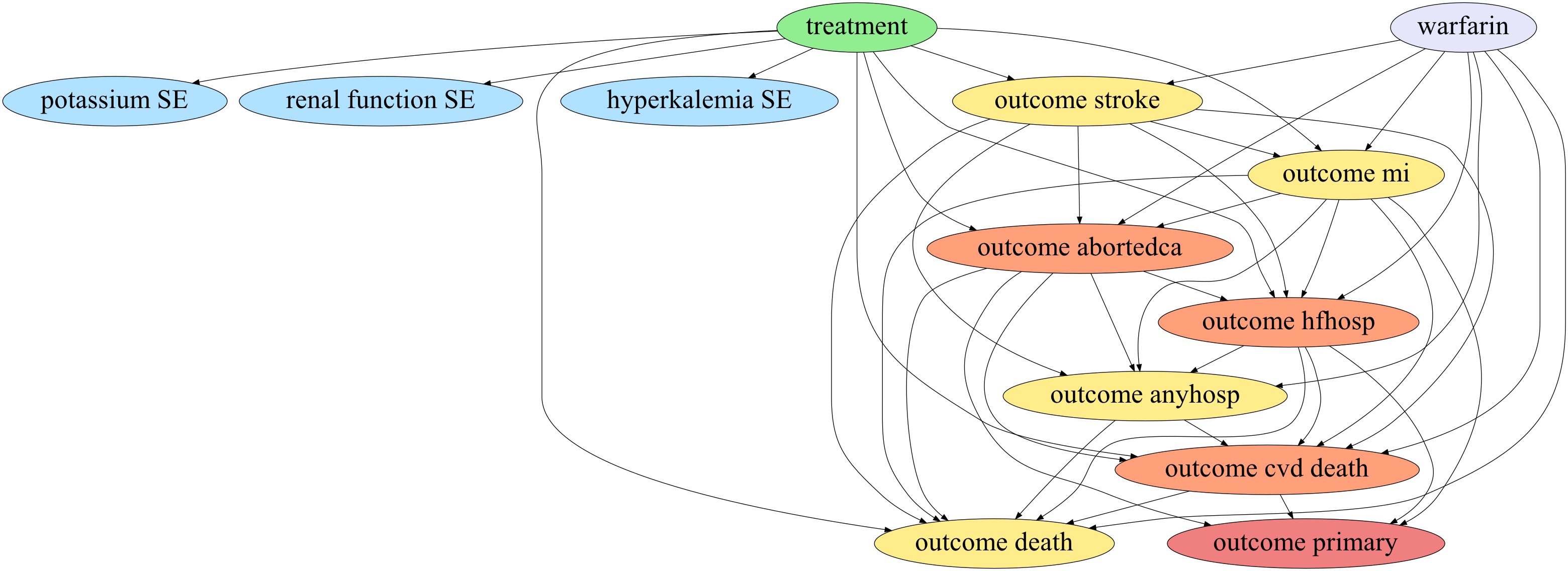}
\caption{DAG for the global dataset - after pruning: \\
non outcome nodes with less than $7$ outbound edges have been pruned, for better visualization.}
\label{fig-DAG-geo-global-pruned}
\end{figure}
\vfill
\end{landscape}

\clearpage
\vfill
\vfill

\clearpage
\begin{landscape}
~
\vfill
\begin{figure}[ht!]\setlength\abovecaptionskip{1cm}
\centering
\includegraphics[width=\paperwidth]{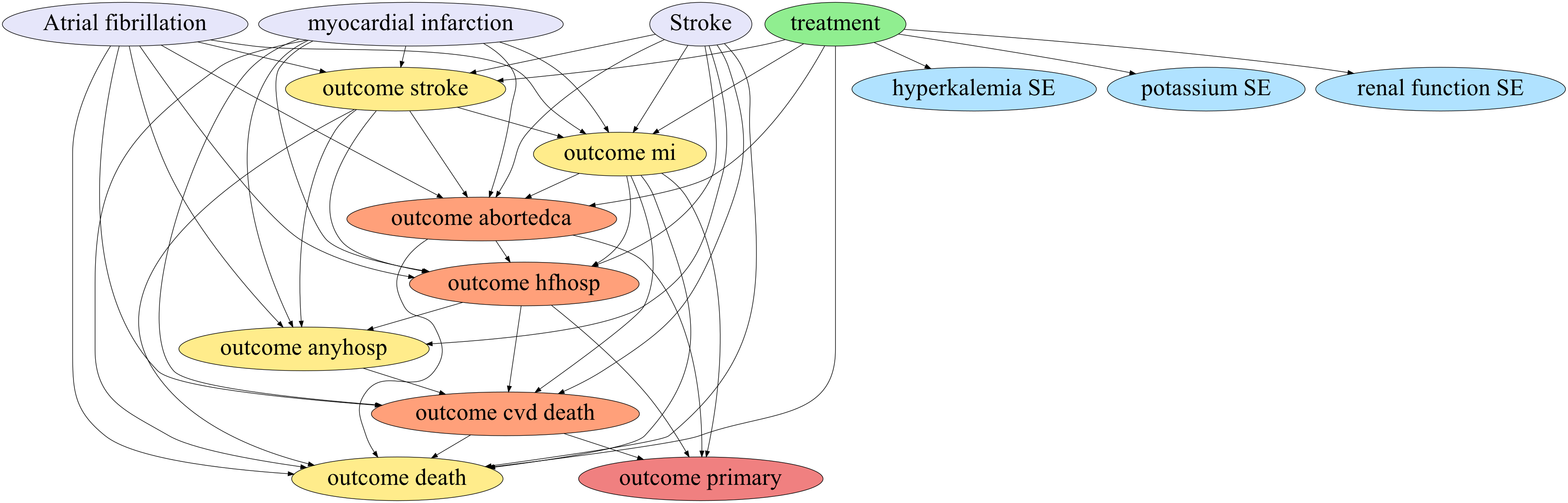}
\caption{DAG for the Western regional subset - after pruning: \\
non outcome nodes with less than $7$ outbound edges have been pruned.}
\label{fig-DAG-geo-regional-west-pruned}
\end{figure}
\vfill
\end{landscape}

\clearpage
\vfill
\vfill

\clearpage
\begin{landscape}
~
\vfill
\begin{figure}[ht!]\setlength\abovecaptionskip{1cm}
\centering
\includegraphics[width=\paperwidth]{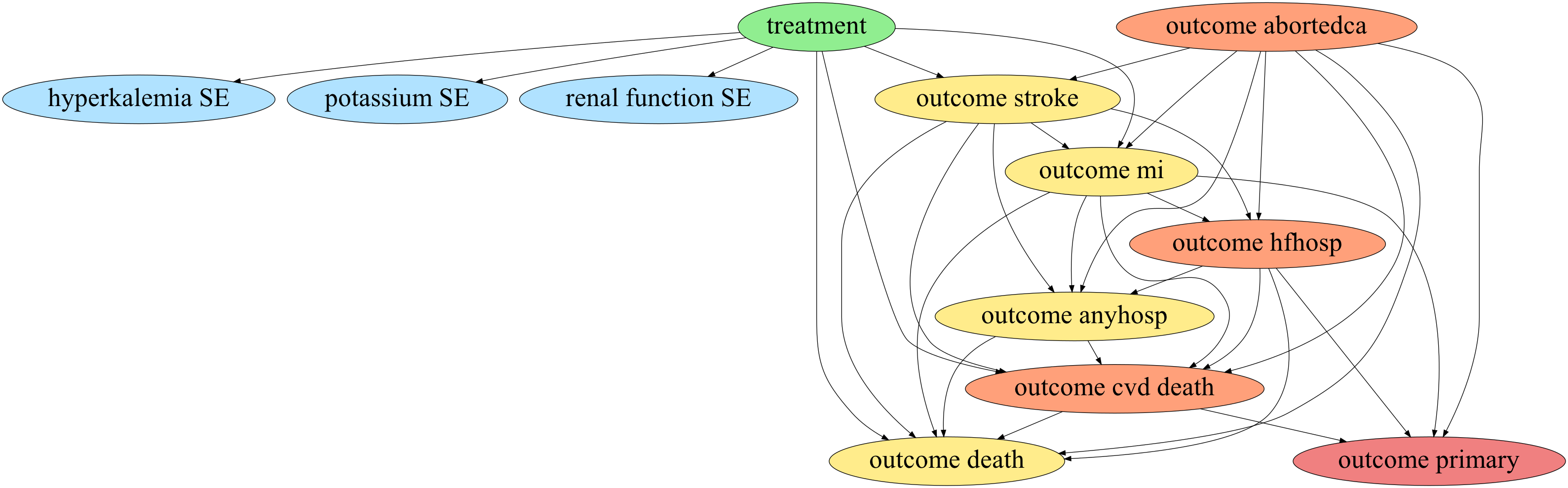}
\caption{DAG for the Eastern regional subset - after pruning: \\
non outcome nodes with less than $7$ outbound edges have been pruned.}
\label{fig-DAG-geo-regional-east-pruned}
\end{figure}
\vfill
\end{landscape}

\end{document}